\documentclass{article}
\usepackage[preprint]{colm2026_conference}
\usepackage{latexsym}
\usepackage[T1]{fontenc}
\usepackage[utf8]{inputenc}
\usepackage{microtype}
\usepackage{graphicx}
\usepackage{amsmath,amssymb}
\usepackage{booktabs,tabularx,array,longtable,multirow}
\usepackage[table]{xcolor}
\usepackage{hyperref}
\usepackage{url}
\usepackage{lineno}
\hypersetup{hidelinks,pdftitle={ProCAP: Decomposed Credit Assignment for Long-Horizon Agentic Reinforcement Learning},pdfauthor={Anonymous authors}}

\newcommand{\procap}{ProCAP}
\newcommand{\ind}{\mathbb{I}}
\newcommand{\clip}{\operatorname{clip}}
\newcommand{\paperdate}{November 27, 2026}
\fancypagestyle{procapfirstpage}{\fancyhead{}\fancyhead[L]{\raisebox{0pt}[0pt][0pt]{\includegraphics[height=17pt]{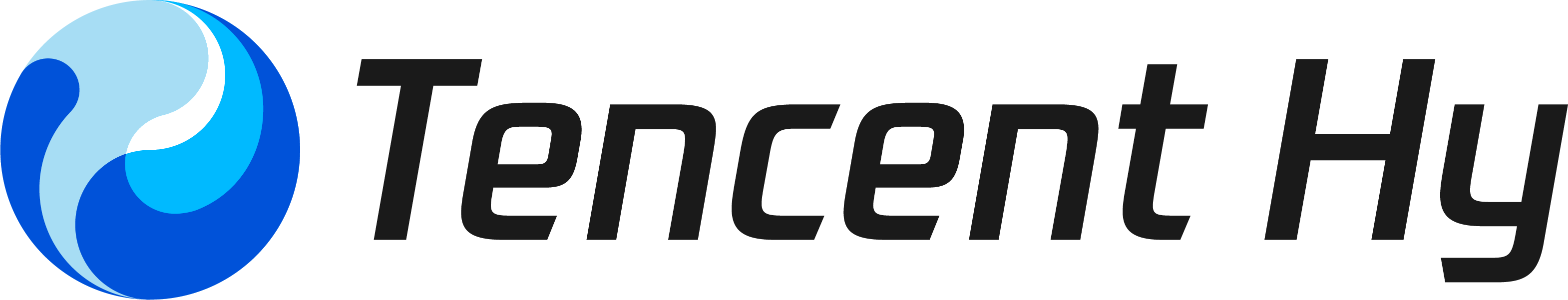}}\strut}\fancyhead[R]{\small\paperdate}}
\title{ProCAP: Decomposing Credit Assignment for\\Long-Horizon Agentic Reinforcement Learning}
\author{Weihong Zhong$^\dagger$\quad Xuan Luo$^\dagger$\quad Mao Zheng$^\dagger$\quad Mingyang Song$^\dagger$\quad \\ \textbf{Xiaocheng Feng$^\ddagger$\quad Bing Qin$^\diamond$}\\
$^\dagger$Tencent Hunyuan\quad  $^\ddagger$Shanghai AI Lab\quad  $^\diamond$Shenzhen Loop Area Institute\\
Email: \{evanwhzhong, ttssxuanluo, moonzheng, nickmysong\}@tencent.com,\\ fengxiaocheng1988@gmail.com, bingqin@slai.edu.cn}
\begin{document}
\raggedbottom
\ifcolmsubmission
\linenumbers
\fi
\maketitle
\lhead{Preprint.}
\thispagestyle{procapfirstpage}
\begin{abstract}
Long-horizon language-model agents must coordinate dozens of local decisions toward a single delayed outcome. While the recent de facto method, Group Relative Policy Optimization (GRPO), avoids a learned critic, it broadcasts one episode-level advantage uniformly across every turn, obscuring which actions helped or harmed the trajectory. Existing turn-level process rewards offer finer supervision, yet reducing each turn to a single score or label blurs a fundamental distinction: an action can advance a subgoal while simultaneously violating core task requirements. To resolve this tension, we introduce \procap{}, a process reward framework that decomposes turn-level credit assignment into multidimensional behavioral assessment and principled, rule-based attribution. Drawing on multidimensional models of human work performance, \procap{} evaluates \emph{Compliance} with task constraints, \emph{Adaptation} to available information, and \emph{Progress} beyond the current state as separate, non-exclusive dimensions. To operationalize these dimensions, \procap{} pairs a frozen LLM judge that assigns categorical turn-level labels from the task and preceding interaction with programmatic checks for execution errors and repeated failures. An explicit attribution layer aggregates these signals into bounded turn credits, directly capping credit whenever Compliance or execution fails, and combines them with GRPO's outcome advantages. Across AppWorld, ClawGym, and PinchBench at 4B and 9B scales, \procap{} yields consistent gains, outperforming GRPO by 4.1 and 9.9 points and the strongest process-reward baseline by 5.9 and 5.4 points in overall average score, respectively.

\end{abstract}
\section{Introduction}
\label{sec:intro}
\begin{figure}[!t]
\centering
\noindent\raisebox{-0.5\height}{\includegraphics[trim=40bp 118bp 64bp 32bp,clip,width=0.60\linewidth]{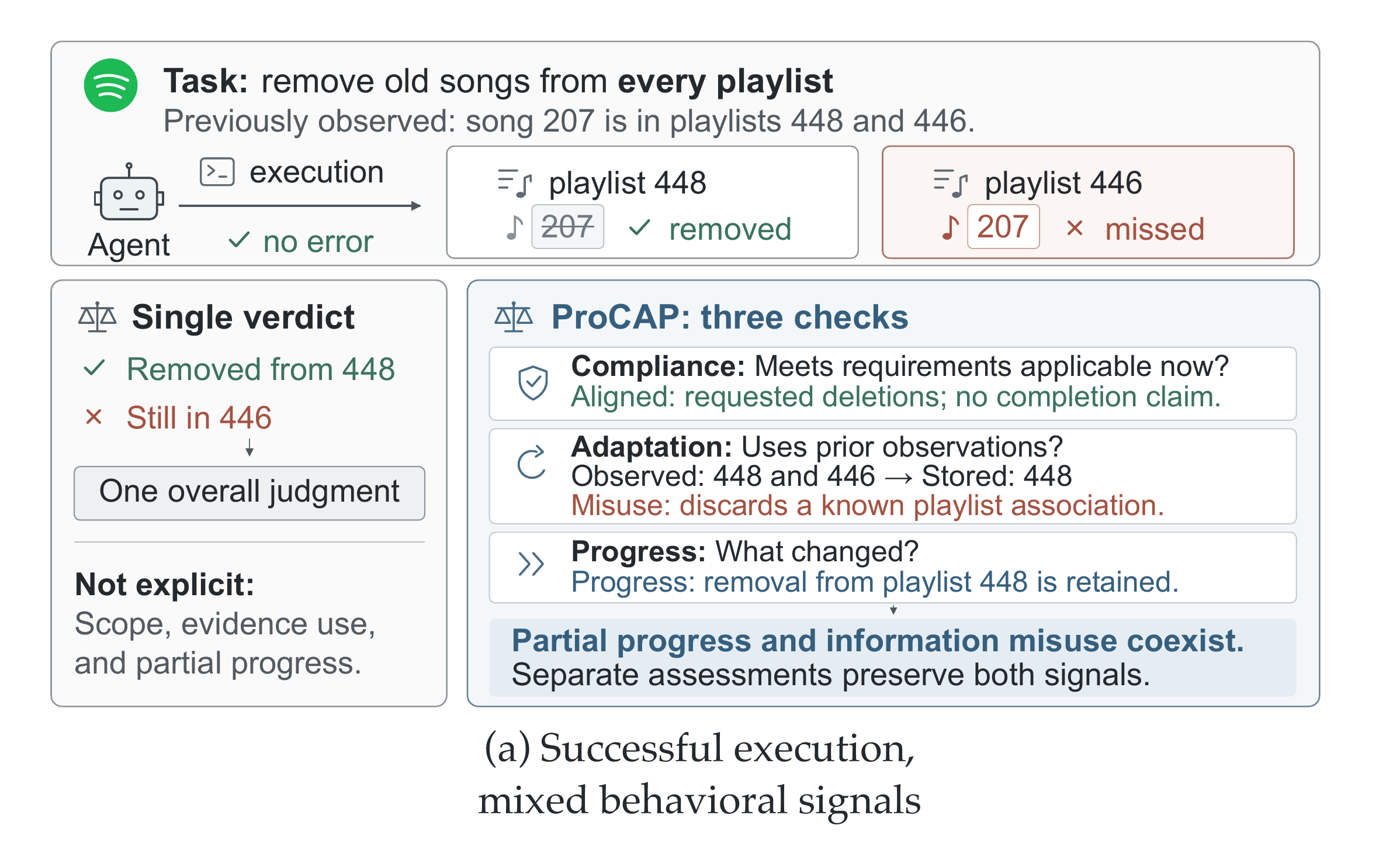}}\hfill\raisebox{-0.5\height}{\includegraphics[width=0.38\linewidth]{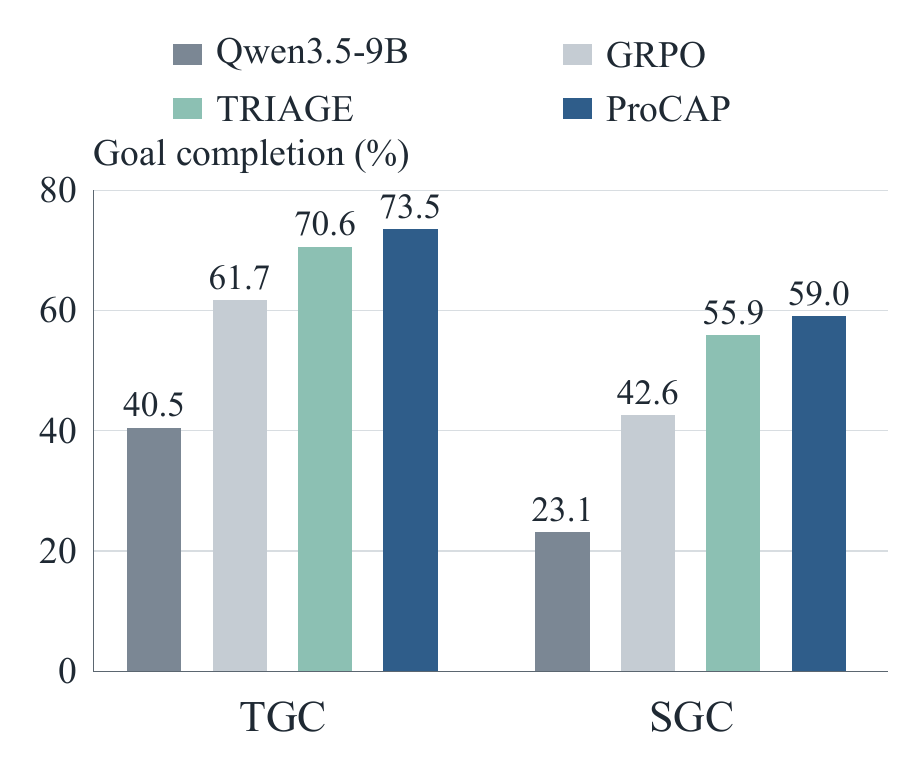}}\par\smallskip
\noindent\begin{minipage}[t]{0.60\linewidth}
\centering\small\bfseries
(a) Partially successful execution,\\mixed behavioral signals
\end{minipage}\hfill\begin{minipage}[t]{0.38\linewidth}
\centering\small\bfseries
(b) Task and scenario\\completion rates
\end{minipage}\par
\caption{(a) Partial progress and information misuse in a playlist-deletion turn. A single-valued song-to-playlist mapping discards one of two observed playlist associations. The deletion succeeds and its result is retained, while the other association remains. Separate assessments preserve both signals. (b) Held-out task (TGC) and scenario (SGC) goal completion for Qwen3.5-9B on AppWorld, pooling normal and challenge splits.}
\label{fig:comparison}
\end{figure}

In long-horizon agentic tasks, a single binary or scalar outcome arrives only after dozens of intermediate tool calls and reasoning steps. Successful trajectories frequently contain redundant or flawed actions, whereas failed trajectories often contain genuine diagnostic progress. Outcome-only methods such as Group Relative Policy Optimization (GRPO) \citep{shao2024deepseekmath} broadcast a single episode-level advantage to every generated token, leaving intra-trajectory credit assignment unresolved---and producing zero outcome gradient whenever all sampled episodes in a group achieve the same final score.

Turn-level process supervision addresses this granularity gap along two main paradigms. Structural and verifier-based methods propagate outcome credit through shared states across rollouts \citep{wang2026g2po,feng2025gigpo} or executable environment checks \citep{li2026vict}, while semantic methods employ LLM judges to evaluate turns in context. Among judge-based approaches, some assign each turn a single exclusive role mapped to a fixed credit \citep{xu2026triage}, others aggregate multiple criteria into scalar scores, gates, or trajectory-level feedback \citep{wang2026cwgrpo,ding2026adarubric,tian2026arco}, and rubric-driven variants operate without turn-level judging \citep{zhang2026trca,gandhi2026draco}. Yet real agent turns are rarely unidimensionally good or bad. In Figure~\ref{fig:comparison}(a), an agent instructed to remove a song from \emph{every} playlist deletes it cleanly from one playlist while overlooking a second playlist it had already observed. The deletion executes successfully and produces retained partial progress, while storing only one playlist per song discards a previously observed playlist association. The same turn therefore provides positive evidence of \emph{Progress} and negative evidence of \emph{Adaptation}.

This example exposes a central design question for process supervision: \emph{which behavioral evidence should be preserved, and how should conflicting evidence govern learning credit?} We argue that turn-level credit assignment should be explicitly \emph{decomposed} along two axes: first, separating \emph{behavioral assessment} (identifying what a turn achieved and where it fell short) from \emph{credit attribution} (deciding how those findings shape policy updates); and second, factoring assessment itself into distinct, non-exclusive behavioral dimensions. Separate assessments preserve both the retained progress and the information-use error; an explicit attribution rule then determines their combined learning credit. A single verdict leaves this trade-off implicit.

Guided by this perspective, we propose \procap{}, a decomposed credit-assignment framework that maintains separate, non-exclusive behavioral assessments and converts them into turn-level credit through a principled, explicit attribution layer. Inspired by multidimensional frameworks of human work performance \citep{koopmans2011frameworks}, a frozen LLM judge assigns task-independent categorical labels along three dimensions: \emph{Compliance} with task requirements and hard boundaries, \emph{Adaptation} to information already revealed in the trajectory, and \emph{Progress} beyond the current state. The judge observes the task, up to five preceding turns, and the target turn with its observation, allowing scoring to proceed online. In parallel, lightweight programmatic detectors flag execution errors and repeated identical failures.

Rather than folding these heterogeneous signals into an opaque scalar, \procap{}'s attribution layer assigns them distinct functional roles governed by explicit design principles. Progress and Adaptation provide signed contributions combined via an idempotent extremum operator that avoids double-counting correlated behavior, whereas execution errors, repeated failures, and Compliance act exclusively as \emph{non-compensatory upper bounds}. Consequently, a core requirement violation strictly caps raw process credit below any positive contribution. The resulting bounded turn credits are centered in exact alignment with GRPO's token-and-episode loss weighting and added to the episode outcome advantage. As previewed in Figure~\ref{fig:comparison}(b), this decomposed formulation yields substantial gains in both task and scenario completion on held-out AppWorld benchmarks.

Our main contributions are threefold:
\begin{itemize}
\item \textbf{Decomposed multidimensional process supervision.} We introduce \procap{}, which decomposes turn-level credit assignment into causal, categorical behavioral assessment (Compliance, Adaptation, Progress), deterministic failure detection, and a separate downstream attribution layer for long-horizon agentic RL.
\item \textbf{Principled non-compensatory attribution.} We design an explicit, task-agnostic attribution layer combining idempotent signed-signal fusion, non-compensatory constraint caps, and GRPO-aligned zero-mean group centering that remains informative even when group outcomes tie.
\item \textbf{Consistent empirical gains across environments and scales.} Evaluated on AppWorld, ClawGym, and out-of-distribution PinchBench at 4B and 9B scales, \procap{} achieves overall averages of 50.4 and 65.1, outperforming GRPO by 4.1 and 9.9 points and the strongest process-reward baseline by 5.9 and 5.4 points, with particularly strong improvements in multi-task scenario completion and cross-benchmark transfer.
\end{itemize}

\section{Related Work}
\subsection{Process Reward Models}
Process supervision provides denser learning signals than terminal outcome rewards by evaluating intermediate reasoning or actions with learned or generative verifiers \citep{uesato2022process,lightman2024verify,wang2024mathshepherd,luo2024omegaprm,zhao2025genprm,khalifa2025thinkprm}.

\paragraph{Learned and structural credit assignment.}
A prominent line of work estimates intermediate value without explicit semantic rubrics. Agent PRMs learn step values or progress directly from trajectories \citep{choudhury2025agentprm,xi2025agentprm,wang2025sparl}, implicit reward formulations derive dense signals from outcome or preference supervision \citep{cui2025prime,liu2025istar}, and potential-based shaping tracks intermediate answer support \citep{wang2025igpo,hu2026siop}. Complementary structural methods propagate episode outcomes across trajectory graphs or shared observation states \citep{li2025salt,wang2026g2po,feng2026rewardflow,yan2026gatedbepo,feng2025gigpo,he2026hgpo}, while hindsight and privileged-information approaches leverage future context for value estimation or segment-level credit \citep{zhou2025sweetrl,tan2026hcapo,lu2026hisr,zhu2026lvrl}. While effective when trajectories share compact state spaces or hindsight is available, structural methods depend on state overlap across rollouts, which can become sparse in open-ended multi-turn environments.

\paragraph{Semantic and executable feedback.}
Judge- and rubric-based methods evaluate actions directly from semantic context \citep{gunjal2025rar,zhu2026coca,liu2026rare}, using dynamic checklists \citep{yin2025dgprm,chae2025webshepherd,ding2026adarubric,tian2026arco}, exclusive role or contribution labels \citep{xu2026triage,wang2026cwgrpo}, or trajectory-level assessments \citep{gandhi2026draco,han2026swetrace}. In parallel, executable checks, oracles, and environment diagnostics supply deterministic supervision \citep{yuan2026vpr,li2026vict,zhang2025rlvmr,jian2026c3,zhang2026trca,li2025rltr}. Most closely related to non-additive gating, MGR \citep{lu2026mgr} decouples local action validity---inferred from syntax checks, execution feedback, and repetition penalties---from global trajectory effectiveness via multiplicative reward gating and a batch-statistics-driven sign-flip schedule, preventing invalid steps in successful rollouts from receiving positive credit. However, its local signal remains purely execution- and syntax-based and cannot distinguish syntactically valid turns that violate task constraints or stall strategically. \procap{} bridges these two directions through a decomposed design: it factors turn evaluation into causal, non-exclusive semantic labels and deterministic failure detectors, then maps them through explicit attribution rules in which both Compliance and execution validity act strictly as non-compensatory caps rather than additive terms. Complementary to RL training, process benchmarks \citep{zheng2025processbench,song2025prmbench,fan2026agentprocessbench} and failure-attribution frameworks \citep{zhang2025whowhen,rafi2026falat} focus on post-hoc evaluation and trajectory diagnosis.

\subsection{Agentic RL}
Interactive benchmarks test multi-turn planning, tool orchestration, and state tracking across web shopping, browser navigation, and rich application ecosystems \citep{yao2022webshop,zhou2024webarena,trivedi2024appworld,bai2026clawgym}. Training language agents in such long-horizon settings has driven rapid progress in multi-turn RL architectures and rollout infrastructures \citep{wang2025ragen,qi2025webrl,xi2025agentgymrl,zhou2024archer,luo2025agentlightning}, as well as specialized methods for multi-step tool invocation \citep{jin2025searchr1,feng2025retool}, milestone discovery \citep{wang2026mira,zheng2026admire}, context folding over extended histories \citep{sun2025foldgrpo}, and turn-level reward shaping \citep{wei2025turnreward}.

Across these diverse rollout pipelines, however, credit assignment over long interaction horizons remains a shared bottleneck whenever supervision is dominated by sparse terminal outcomes. \procap{} is orthogonal to the underlying rollout or context-management architecture: it operates purely on the advantage construction step of group-based policy optimization, making it directly applicable to existing multi-turn RL systems.

\section{Preliminaries}
\label{sec:prelim}
We consider a language-model agent solving a task $x$ through multi-turn interaction. At turn $t$, the policy $\pi_\theta$ conditions on the interaction history $h_t$ and generates a turn $a_t$ comprising optional reasoning and an action submitted to the environment, which returns an observation $o_t$. An episode $\tau=((a_t,o_t))_{t=1}^{T}$ receives a single terminal outcome reward $R\in[0,1]$ upon completion and no intermediate environment reward.

GRPO \citep{shao2024deepseekmath} eliminates a learned critic by sampling a group of $G$ independent episodes for the same task $x$ and normalizing each episode's outcome against its group. Indexing episodes by $i\in\{1,\dots,G\}$, let $a_{it}$, $o_{it}$, $T_i$, and $R_i$ denote the components and outcome of episode $i$. Its outcome advantage is
\begin{equation}
A_i^{\mathrm{out}}=\frac{R_i-\bar R_x}{\sigma_x+\delta},
\label{eq:outcome}
\end{equation}
where $\bar R_x=G^{-1}\sum_{j=1}^{G}R_j$, $\sigma_x^2=(G-1)^{-1}\sum_{j=1}^{G}(R_j-\bar R_x)^2$ is the sample variance, and $\delta=10^{-6}$ (we use $G=8$ throughout). Every generated token in episode $i$ receives the same advantage $A_i^{\mathrm{out}}$. This introduces two limitations in long-horizon settings: (i) intra-episode credit assignment is completely flat, rewarding harmful turns in successful episodes and penalizing useful turns in failed ones; and (ii) whenever all $G$ episodes share the same outcome, $A_i^{\mathrm{out}}=0$, leaving no outcome-driven policy gradient beyond KL regularization.

Let $n_{it}$ denote the number of generated tokens in turn $t$ of episode $i$, and $N_i=\sum_{t=1}^{T_i}n_{it}$ the total number of loss-bearing generated tokens in episode $i$ (excluding prompts and observations). To provide fine-grained supervision without distorting episode-level optimization, a turn-level process signal should be locally computable, preserve episode outcome rewards $R_i$, and remain zero-mean under GRPO's loss weighting. Section~\ref{sec:method} constructs such a turn-level credit and integrates it with $A_i^{\mathrm{out}}$.

\section{Method}
\label{sec:method}
\begin{figure}[!t]
\centering
\includegraphics[width=\textwidth]{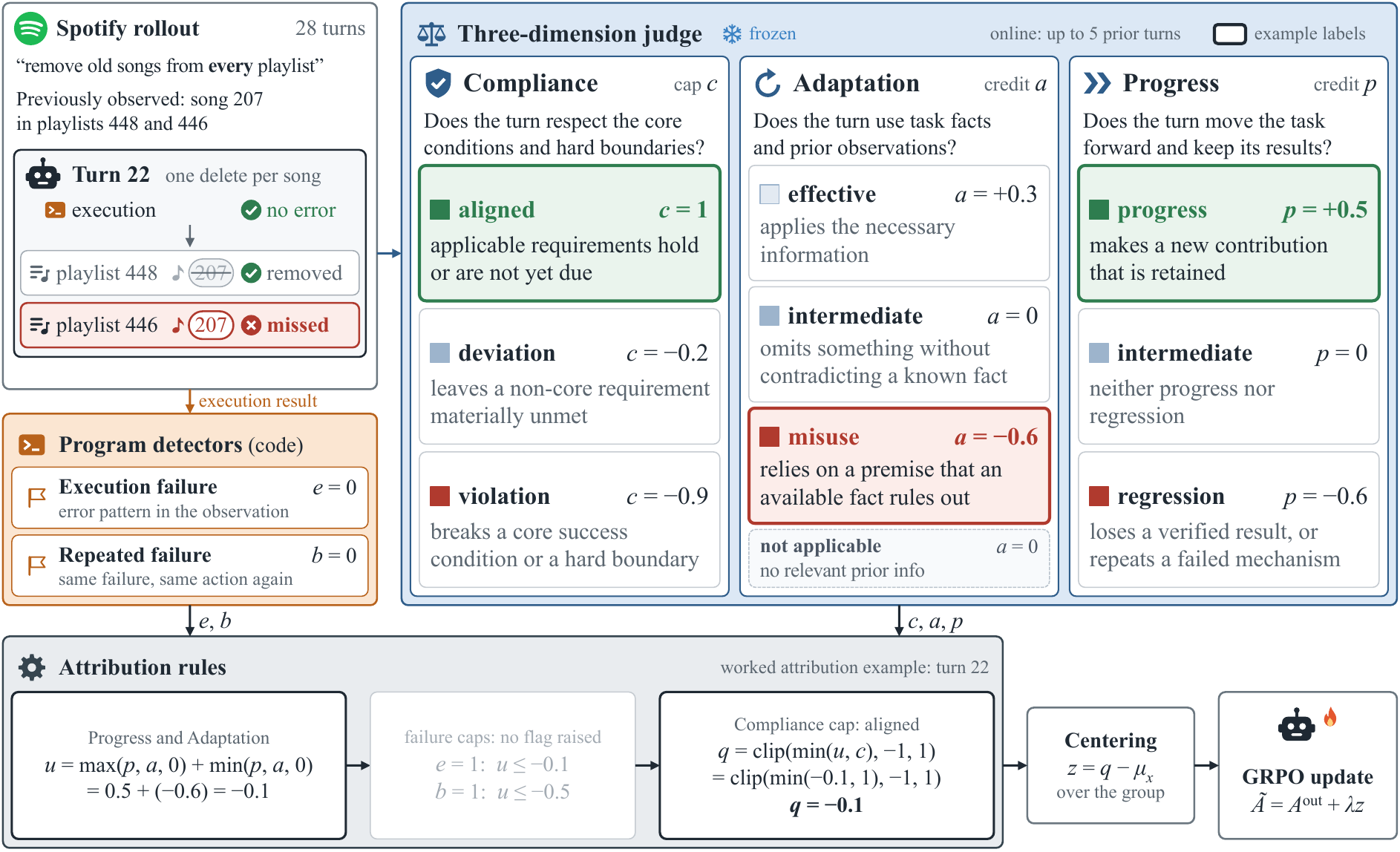}
\caption{Overview of \procap{}. A frozen causal judge assigns categorical labels for Compliance, Adaptation, and Progress; programmatic detectors flag execution and repetition failures. Explicit rules fuse these signals, apply caps, and center credits for GRPO.}
\label{fig:method}
\end{figure}

Because only the terminal outcome verifies whether task $x$ was solved, \procap{} retains GRPO's outcome advantage (Equation~\ref{eq:outcome}) and augments it with a bounded, group-centered turn credit. Our design rests on a single architectural principle: \emph{decomposing credit assignment into multidimensional behavioral assessment and explicit credit attribution}. Every agent turn simultaneously faces three distinct behavioral tests---\emph{Compliance} with task requirements and boundaries, \emph{Adaptation} to information already revealed in the trajectory, and \emph{Progress} beyond the current state---alongside deterministic execution validity. Collapsing these tests into one overall verdict forces a judge to weigh gains against errors implicitly. \procap{} instead preserves separate, non-exclusive categorical labels and deterministic error flags, delegating all trade-offs to an explicit, task-agnostic attribution layer.

Figure~\ref{fig:method} summarizes the four stages of the pipeline: Section~\ref{sec:judge} describes the causal three-dimension judge, Section~\ref{sec:detectors} specifies the programmatic failure detectors, Section~\ref{sec:rules} formalizes the attribution rules and objective-aligned group centering, and Section~\ref{sec:optimization} integrates the resulting process term into GRPO.

\subsection{Three-Dimension Judge}
\label{sec:judge}
\paragraph{Multidimensional decomposition.}
Our assessment taxonomy is grounded in the multidimensional structure of human work performance \citep{koopmans2011frameworks}, which distinguishes core task accomplishment from contextual rule adherence \citep{borman1997task} and adaptive responses to emerging information \citep{pulakos2000adaptability}. Translating this view to multi-turn agents yields three orthogonal, non-exclusive dimensions:
\begin{itemize}
\item \emph{Compliance} evaluates adherence to the task's core success conditions and hard boundaries (e.g., requested scope, authorization, truthfulness, and permitted sources): \emph{violation} if an action, plan, claim, or delivery breaches a core boundary; \emph{deviation} if only a non-core requirement is materially unmet; and \emph{aligned} otherwise. Requirements are assessed when they apply: action boundaries at the attempt, prerequisites before dependent actions, and delivery correctness at submission or when explicitly due.
\item \emph{Adaptation} evaluates how faithfully the turn incorporates facts already revealed in the trajectory: \emph{misuse} if the action relies on a premise contradicted by available evidence; \emph{effective} if relevant prior information is properly applied; \emph{intermediate} if helpful context is omitted without direct contradiction; or \emph{not applicable} (mapped to zero credit) when no prior evidence governs the step.
\item \emph{Progress} evaluates net forward movement beyond the starting state of the turn: \emph{regression} if a verified result is lost or a failed mechanism is repeated without substantive change; \emph{progress} if the turn achieves a retained new contribution; and \emph{intermediate} otherwise.
\end{itemize}
Evaluating these dimensions independently disentangles distinct failure modes: a productive step can breach task scope, a compliant step can stall, and an action that properly uses prior observations can still trigger an unforeseen execution error.

\paragraph{Scale-free categorical judging.}
For each target turn $(i,t)$, the frozen LLM judge returns a validity status and, when scorable, a triplet of categorical labels rather than continuous numeric scores:
\begin{equation}
\boldsymbol\ell_{it}=(\ell_{it}^{\mathrm C},\ell_{it}^{\mathrm A},\ell_{it}^{\mathrm P}).
\label{eq:scores}
\end{equation}
Restricting the judge to a compact, task-independent categorical rubric avoids entangling semantic recognition with arbitrary numerical calibration across diverse environments, keeping the downstream credit mapping transparent and auditable. Following a fixed decision order, the judge tests the severe condition first and the positive condition next; when context is insufficient to judge, it emits an unscorable status (triggering the group fallback in Section~\ref{sec:rules}) rather than defaulting to an intermediate label.

\paragraph{Causal context window.}
The rubric is shared across all tasks, with environment-specific facts provided in the prompt. For each target turn, the judge receives the task specification, up to five preceding turns (truncated while preserving policy reasoning), and the complete target turn together with its observation $o_{it}$. In our default configuration, no future turns are shown, so assessment can execute online once $o_{it}$ arrives (we compare causal and hindsight windows in Section~\ref{sec:analysis}). Appendix~\ref{app:rubric} details the full rubric and judge interface.

\subsection{Program Detectors}
\label{sec:detectors}
Semantic assessment alone can miss low-level runtime errors, while an action that reasons soundly over prior observations may still fail upon execution. To ground credit assignment in observable environment outcomes, \procap{} pairs the semantic judge with deterministic programmatic detectors (Figure~\ref{fig:method}) that flag execution errors and repeated identical failures.

\paragraph{Execution failure detection.}
For text observations, the detector scans $o_{it}$ using an ordered, case-insensitive list of regular expressions (Appendix~\ref{app:detectors}). The first matching category determines the failure type $f_{it}$ and sets $e_{it}=1$; if no pattern matches, $f_{it}=\varnothing$ and $e_{it}=0$. Priority ordering ensures that specific diagnostics (e.g., \texttt{TypeError}) take precedence over generic patterns (e.g., \texttt{traceback}) regardless of text position, and multiple matches within an observation yield a single turn-level flag rather than an error count.

\paragraph{Repeated failure detection.}
To penalize unproductive stalling loops without discouraging deliberate debugging, let $v(a)$ normalize a generated turn by collapsing whitespace runs to single spaces and trimming boundaries while retaining policy reasoning. A repeated failure is flagged when two consecutive turns produce both the same failure category and the identical normalized output: $b_{i1}=0$ and, for $t>1$,
\begin{equation}
 b_{it} = e_{it}\ind\{f_{it}=f_{i,t-1}\}\,\ind\{v(a_{it})=v(a_{i,t-1})\}.
\label{eq:repeat}
\end{equation}
Retaining reasoning in $v(a)$ intentionally distinguishes mechanical repetition from informed retry: an agent that adjusts its diagnosis and re-invokes an API under a revised hypothesis does not trigger $b_{it}=1$. Optional structured environment adapters follow the same interface (Appendix~\ref{app:detectors}).

Programmatic flags and semantic labels are strictly complementary. In Section~\ref{sec:rules}, $e_{it}$ and $b_{it}$ impose deterministic upper bounds on raw credit so that positive semantic labels cannot override a runtime crash, while semantic labels capture misuse of prior information despite successful execution (Figure~\ref{fig:comparison}(a)), which raises neither execution flag.

\subsection{Attribution Rules and Centering}
\label{sec:rules}
\begin{table}[!t]
\centering\small
\begin{tabularx}{\textwidth}{@{}ll>{\raggedright\arraybackslash}Xlr@{}}
\toprule
Dimension & Role & Criterion & Judge label & Value \\
\midrule
\multirow[t]{3}{*}{Progress ($p$)} & \multirow[t]{3}{*}{Signed credit} & \multirow[t]{3}{=}{Net retained contribution beyond the turn's starting state.} & progress & $+0.5$ \\
& & & intermediate & $0$ \\
& & & regression & $-0.6$ \\
\midrule
\multirow[t]{4}{*}{Adaptation ($a$)} & \multirow[t]{4}{*}{Signed credit} & \multirow[t]{4}{=}{Faithful use of task facts and prior trajectory observations.} & effective & $+0.3$ \\
& & & intermediate & $0$ \\
& & & misuse & $-0.6$ \\
& & & not applicable & $0$ \\
\midrule
\multirow[t]{3}{*}{Compliance ($c$)} & \multirow[t]{3}{*}{Upper-bound cap} & \multirow[t]{3}{=}{Adherence to core task requirements, scope, and hard boundaries.} & aligned & $1$ \\
& & & deviation & $-0.2$ \\
& & & violation & $-0.9$ \\
\bottomrule
\end{tabularx}
\caption{Categorical label-to-value mapping of \procap{}. Progress and Adaptation supply signed credits fused via Equation~\ref{eq:credit}, whereas Compliance acts strictly as a non-compensatory upper bound (Equation~\ref{eq:gates}).}
\label{tab:mapping}
\end{table}

The attribution layer converts the categorical triplet $\boldsymbol\ell_{it}$ and deterministic flags $(e_{it},b_{it})$ into a bounded raw turn credit $q_{it}$ before group centering. Rather than tuning ad-hoc per-environment reward weights, the layer is governed by three explicit design principles: \emph{non-double-counting of correlated signals}, \emph{error dominance}, and \emph{non-compensatory constraint satisfaction}.

\paragraph{Idempotent signal fusion and priority ordering.}
Progress and Adaptation supply signed behavioral contributions (Table~\ref{tab:mapping}): we assign $p=+0.5$ for \emph{progress}, $a=+0.3$ for \emph{effective} Adaptation, $0$ for \emph{intermediate} labels or \emph{not applicable} Adaptation, and $p=a=-0.6$ for \emph{regression} or \emph{misuse}. This assignment satisfies the strict priority ordering $|p_-|=|a_-| > p_+ > a_+ > 0$ ($0.6 > 0.5 > 0.3$): advancing the task state earns more than merely applying existing information without new progress, while a severe error in either dimension outweighs the maximum positive contribution of the other.

Because Progress and Adaptation are non-exclusive, additive aggregation ($p+a$) would double-count correlated behaviors and inflate credit variance. We instead combine them with an idempotent signed-extremum operator that retains only the strongest positive and strongest negative term:
\begin{equation}
 u=\max(p,a,0)+\min(p,a,0).
\label{eq:credit}
\end{equation}
For same-sign inputs, Equation~\ref{eq:credit} collapses to the dominant magnitude---progress paired with effective information use receives $0.5$ (not $0.8$), and regression paired with misuse receives $-0.6$ (not $-1.2$). For mixed-sign inputs, both terms remain active. In the playlist example (Figure~\ref{fig:comparison}(a)), the retained deletion contributes $p=+0.5$, while discarding a known playlist association contributes $a=-0.6$, yielding $u=-0.1$. This preserves the information-use penalty while crediting the useful work.

\paragraph{Non-compensatory execution and compliance ceilings.}
Execution validity and task Compliance represent hard constraints rather than tradable bonuses. Accordingly, both enter credit assignment exclusively as upper-bound caps rather than additive penalties. Detected execution failures and repeated identical failures impose deterministic ceilings:
\begin{equation}
\begin{aligned}
 u&\leftarrow\min(u,-0.1)\quad\text{if }e=1,\\
 u&\leftarrow\min(u,-0.5)\quad\text{if }b=1.
\end{aligned}
\label{eq:caps}
\end{equation}
Because these are upper bounds, an execution error reduces $u=0.5$ to $-0.1$ while leaving a more severe semantic credit ($u=-0.6$) unchanged.

Compliance acts under the same non-compensatory principle. Setting $c=1$ for \emph{aligned}, $c=-0.2$ for \emph{deviation}, and $c=-0.9$ for \emph{violation}, the raw turn credit is
\begin{equation}
 q=\clip(\min(u,c),-1,1).
\label{eq:gates}
\end{equation}
Here $c=1$ acts as an identity element (never adding an unearned bonus), a non-core deviation prevents positive raw credit, and a core violation fixes $q=-0.9$---strictly below every possible fused credit---so neither progress nor effective adaptation can compensate for breaching a core requirement. Across failure modes, penalty severity scales cleanly from a single execution error ($-0.1$) and non-core deviation ($-0.2$), to repeated execution failure ($-0.5$), semantic regression or misuse ($-0.6$), and core violation ($-0.9$). We keep this single $[-0.9,0.5]$ instantiation fixed across all benchmarks and model scales, controlling overall process strength solely through $\lambda$ in Section~\ref{sec:optimization}. In Figure~\ref{fig:method}, the playlist example has aligned Compliance ($c=1$) and no execution failures ($e=b=0$), so the fused credit passes through the caps unchanged: $q=-0.1$.

\paragraph{Objective-aligned group centering.}
Let $\mathcal K_{it}$ denote the $n_{it}$ loss-bearing generated token positions of turn $t$ in episode $i$. A turn is valid when Compliance and Progress are scored and Adaptation is either scored or not applicable. Let $\chi_x=1$ if process shaping is active and every episode in task group $x$ passes judge-coverage and token-span validation, and $\chi_x=0$ otherwise. For valid groups, we compute episode-level token-weighted means and center equally across the $G$ episodes:
\begin{align}
 \bar q_i&=\frac{1}{N_i}\sum_{t=1}^{T_i}n_{it}q_{it},
\label{eq:episodemean}\\
 \mu_x&=\frac1G\sum_{i=1}^{G}\bar q_i,
\label{eq:baseline}
\end{align}
assigning each token $k\in\mathcal K_{it}$ the centered process advantage
\begin{equation}
 z_{ik}=\begin{cases}
 q_{it}-\mu_x,&\chi_x=1,\\
 0,&\chi_x=0.
 \end{cases}
\label{eq:centering}
\end{equation}
Notice that Equations~\ref{eq:episodemean}--\ref{eq:centering} mirror the exact token-and-episode weighting of the GRPO objective (Equation~\ref{eq:objective}), guaranteeing zero-mean alignment under the training loss measure:
\begin{equation}
\frac1G\sum_{i=1}^{G}\frac1{N_i}\sum_{k=1}^{N_i} z_{ik}=0.
\label{eq:zeromean}
\end{equation}
Process shaping thus redistributes relative credit across turns without shifting the group's net baseline or biasing updates toward longer trajectories. Each centered credit is broadcast directly to all tokens in its span $\mathcal K_{it}$; if any turn fails validation ($\chi_x=0$), the group cleanly falls back to standard outcome-only GRPO.

\subsection{Policy Optimization}
\label{sec:optimization}
\procap{} combines the centered turn credit with GRPO's episode-level outcome advantage at every generated token $k\in\mathcal K_{it}$:
\begin{equation}
 \widetilde A_{ik}=A_i^{\mathrm{out}}+\lambda z_{ik},
\label{eq:advantage}
\end{equation}
with $\lambda=1$ by default.

Let $y_{ik}$ ($k=1,\ldots,N_i$) denote the loss-bearing generated tokens of episode $i$ (excluding prompts and observations), and let $r_{ik}=\pi_\theta(y_{ik})/\pi_{\mathrm{old}}(y_{ik})$ be the importance ratio conditioned on the full interaction prefix. For a task group $x$, we maximize the clipped GRPO objective
\begin{equation}
 \mathcal J(\theta)=\frac1G\sum_{i=1}^{G}\frac1{N_i}\sum_{k=1}^{N_i}\Bigl[\min\!\bigl(r_{ik}\widetilde A_{ik},\,\clip(r_{ik},1-\varepsilon,1+\varepsilon)\,\widetilde A_{ik}\bigr)-\beta\widehat D_{\mathrm{KL},ik}\Bigr],
\label{eq:objective}
\end{equation}
where $\varepsilon$ controls ratio clipping and $\beta$ weights the tokenwise KL penalty toward the reference policy $\pi_{\mathrm{ref}}$. Losses are averaged over tokens within each episode, equally across episodes, and then across tasks in a batch.

This zero-mean advantage shaping yields two useful optimization properties. First, it naturally mitigates GRPO's \emph{zero-advantage collapse}: whenever all $G$ episodes in a group obtain identical outcomes ($A_i^{\mathrm{out}}=0$), differing turn credits still supply informative policy-gradient updates for $\lambda>0$. Second, because credit is relative to the group mean $\mu_x$, even a modestly positive raw credit ($q_{it}>0$) becomes negative if it falls below the group average, while uniform credits, failed validation ($\chi_x=0$), or $\lambda=0$ recover standard GRPO exactly.

\section{Experiments}
\label{sec:experiments}
\subsection{Setup}
\paragraph{Benchmarks and metrics.}
We evaluate on three multi-turn agent benchmarks spanning diverse tool ecosystems and out-of-distribution (OOD) generalization. \textbf{AppWorld} \citep{trivedi2024appworld} comprises 168 normal tasks across 56 scenarios and 417 challenge tasks across 139 scenarios; we report task goal completion (TGC) and scenario goal completion (SGC), where SGC requires every task within a scenario to succeed, and pool both splits for the AppWorld average. \textbf{ClawGym-Bench} \citep{bai2026clawgym} contains 200 tasks spanning Office, Tech, and Analysis domains; its Hybrid metric weights programmatic checks at 0.7 and rubric scores at 0.3 when both apply, and uses the code score alone for code-only tasks. \textbf{PinchBench} \citep{pinchbench2026leaderboard} contains 147 tasks held out entirely from training to assess cross-benchmark OOD transfer. The overall average weights pooled AppWorld TGC, pooled AppWorld SGC, ClawGym All, and PinchBench equally.

\paragraph{Training and judge configuration.}
We fine-tune official post-trained Qwen3.5-4B and Qwen3.5-9B models separately on AppWorld's 90 training tasks and ClawGym's balanced 2,000-task training set; PinchBench is evaluated zero-shot using the corresponding ClawGym-trained checkpoint for each method and model scale. All runs train for 100 steps on 16 GPUs across two nodes, sampling 16 tasks per step with $G=8$ episodes per task, followed by a single policy update with a batch size of 128, learning rate $10^{-6}$, ratio clipping $\varepsilon=0.2$, and KL coefficient $\beta=0.04$.

For process supervision, \procap{} uses a frozen Qwen3.8-27B judge with thinking disabled and temperature 0, applying the task-independent categorical rubric in Appendix~\ref{app:rubric} over a causal 5+0 context window of up to five preceding turns and no future turns, with $\lambda=1$.

\paragraph{Baselines.}
We compare \procap{} against outcome-only RL and two representative turn-level credit-assignment paradigms:
\begin{itemize}
\item \textbf{GRPO} \citep{shao2024deepseekmath}: the outcome-only baseline assigning the episode advantage of Equation~\ref{eq:outcome} uniformly to all tokens.
\item \textbf{G2PO} \citep{wang2026g2po}: a judge-free structural method that merges the $G=8$ episodes of a task into a state-transition graph over identical observations, values each state by discounted mean episode outcomes with $\gamma=0.95$, and adds relative state-value advantages to the episode advantage with weight 1.
\item \textbf{TRIAGE} \citep{xu2026triage}: a semantic judge method that assigns each turn one exclusive role among decisive, exploration, no-progress, and regression, mapped to fixed credits $(1,0.5,-0.1,-0.5)$, added to GRPO with weight 0.2 and batch-whitened. We implement TRIAGE following the paper's equations using the same frozen Qwen3.8-27B judge with thinking disabled and five turns of context on each side of the target turn.
\end{itemize}

\subsection{Main Results}
\begin{table}[!t]
\centering\small
\setlength{\tabcolsep}{1.5pt}
\newcommand{\maintworow}[1]{\smash{\raisebox{-0.5\dimexpr\normalbaselineskip+\aboverulesep+\cmidrulewidth+\belowrulesep\relax}{#1}}}
\newcommand{\mainthreerow}[1]{\smash{\raisebox{-\dimexpr\normalbaselineskip+\aboverulesep+\cmidrulewidth+\belowrulesep\relax}{#1}}}
\begin{tabular*}{\linewidth}{@{\extracolsep{\fill}}l*{12}{c}@{}}
\toprule
\multicolumn{1}{c}{\mainthreerow{Method}} & \multicolumn{6}{c}{AppWorld} & \multicolumn{4}{c}{\maintworow{ClawGym}} & \multicolumn{1}{c}{\maintworow{PinchBench}} & \multicolumn{1}{c}{\mainthreerow{Avg.}}\\
\cmidrule(lr){2-7}
& \multicolumn{2}{c}{Normal} & \multicolumn{2}{c}{Challenge} & \multicolumn{2}{c}{Avg.} & \multicolumn{4}{c}{} & & \\
\cmidrule(lr){2-3}\cmidrule(lr){4-5}\cmidrule(lr){6-7}\cmidrule(lr){8-11}\cmidrule(lr){12-12}
& TGC & SGC & TGC & SGC & TGC & SGC & Office & Tech & Analysis & All & \multicolumn{1}{c}{All} & \\
\midrule\multicolumn{13}{l}{\emph{Qwen3.5-4B}}\\
Initial model & 23.8 & 7.1 & 14.1 & 2.2 & 16.9 & 3.6 & 54.3 & 48.1 & 51.3 & 51.3 & 51.6 & 30.9 \\
GRPO & \textbf{68.5} & 48.2 & 41.7 & 22.3 & 49.4 & 29.7 & 58.5 & 48.2 & \textbf{60.6} & 55.7 & 50.2 & 46.3 \\
G2PO & 62.5 & 42.9 & 31.6 & 17.3 & 40.5 & 24.6 & 53.4 & 41.0 & 47.1 & 47.3 & 49.6 & 40.5 \\
TRIAGE & 61.9 & 42.9 & 42.7 & 23.0 & 48.2 & 28.7 & 52.7 & 43.9 & 53.2 & 49.9 & 51.2 & 44.5 \\
\textbf{\procap{}} & 65.5 & \textbf{50.0} & \textbf{45.8} & \textbf{33.1} & \textbf{51.5} & \textbf{37.9} & \textbf{60.8} & \textbf{53.9} & 56.9 & \textbf{57.3} & \textbf{55.0} & \textbf{50.4} \\
\midrule\multicolumn{13}{l}{\emph{Qwen3.5-9B}}\\
Initial model & 60.7 & 42.9 & 32.4 & 15.1 & 40.5 & 23.1 & 59.1 & 56.8 & 57.7 & 57.9 & 55.2 & 44.2 \\
GRPO & 73.2 & 53.6 & 57.1 & 38.1 & 61.7 & 42.6 & 65.6 & 57.5 & 63.5 & 62.2 & 54.2 & 55.2 \\
G2PO & 69.6 & 55.4 & 51.8 & 30.2 & 56.9 & 37.4 & 64.4 & 55.6 & 64.2 & 61.4 & 59.0 & 53.7 \\
TRIAGE & 80.4 & 66.1 & 66.7 & 51.8 & 70.6 & 55.9 & 57.6 & 54.5 & 61.6 & 57.8 & 54.4 & 59.7 \\
\textbf{\procap{}} & \textbf{83.3} & \textbf{75.0} & \textbf{69.5} & \textbf{52.5} & \textbf{73.5} & \textbf{59.0} & \textbf{68.3} & \textbf{62.2} & \textbf{67.9} & \textbf{66.1} & \textbf{61.7} & \textbf{65.1} \\
\bottomrule
\end{tabular*}
\caption{Main evaluation results (\%) across AppWorld, ClawGym, and out-of-distribution PinchBench. AppWorld Avg.\ pools normal and challenge splits; overall Avg.\ weights AppWorld TGC, AppWorld SGC, ClawGym All, and PinchBench equally. Bold marks the best score per column within each model scale.}
\label{tab:main}
\end{table}

Table~\ref{tab:main} summarizes performance across all three benchmarks and both model scales. \procap{} achieves the highest overall average at both scales, reaching 50.4 at 4B and 65.1 at 9B. It outperforms GRPO by 4.1 and 9.9 points and the strongest process-reward baseline by 5.9 and 5.4 points, respectively. Across the three benchmarks, these gains reflect three distinct behavioral strengths.

\paragraph{Scenario-level completeness on AppWorld.}
While \procap{} improves TGC over GRPO by 2.1 points at 4B and 11.8 points at 9B, its largest relative advantage appears in SGC, where it reaches 37.9 versus 29.7 at 4B and 59.0 versus 42.6 at 9B, outperforming GRPO by 8.2 and 16.4 points. At 4B, \procap{} solves 12 more individual tasks than GRPO yet completes 16 more multi-task scenarios. Because SGC requires every task in a scenario to pass all checks, even a single overlooked scope constraint prevents scenario completion. Penalizing core violations via a non-compensatory Compliance cap directly targets these near-miss failures, and these gains carry over cleanly to the harder challenge split, where \procap{} leads GRPO in SGC by 10.8 points at 4B and 14.4 points at 9B.

\paragraph{Consistent cross-domain gains on ClawGym.}
ClawGym tests heterogeneous workflows across Office, Tech, and Analysis. At 4B, \procap{} leads on Office at 60.8 and Tech at 53.9, reaching 57.3 overall compared to 55.7 for GRPO. At 9B, it leads across all three domains with 68.3 on Office, 62.2 on Tech, and 67.9 on Analysis, reaching 66.1 overall and surpassing GRPO by 3.9 points. Whereas different process-reward paradigms exhibit distinct strengths across environments, \procap{} maintains consistent improvements across both AppWorld and ClawGym as model capacity scales.

\paragraph{Out-of-distribution transfer on PinchBench.}
PinchBench evaluates whether process-shaped ClawGym policies generalize to unseen tasks. At 4B, \procap{} is the only method that improves over the initial model, gaining 3.4 points to reach 55.0, whereas outcome-only GRPO gains 4.4 points in-domain on ClawGym yet drops 1.4 points on PinchBench. At 9B, \procap{} again achieves the highest OOD score at 61.7, outperforming the initial model by 6.5 points and GRPO by 7.5 points. Together, these results indicate that penalizing scope violations, premise misuse, and repeated execution failures encourages generalizable problem-solving habits rather than benchmark-specific shortcuts.

\subsection{Analysis}
\label{sec:analysis}
\label{sec:ablation}
\begin{table}[!t]
\centering\small
\setlength{\tabcolsep}{0pt}
\newcommand{\abltworow}[1]{\smash{\raisebox{-0.5\dimexpr\normalbaselineskip+\aboverulesep+\cmidrulewidth+\belowrulesep\relax}{#1}}}
\newcommand{\ablthreerow}[1]{\smash{\raisebox{-\dimexpr\normalbaselineskip+\aboverulesep+\cmidrulewidth+\belowrulesep\relax}{#1}}}
\begin{tabular*}{\linewidth}{@{\extracolsep{\fill}}ccrrrrrr@{}}
\toprule
\multicolumn{1}{c}{\ablthreerow{Window}} & \multicolumn{1}{c}{\ablthreerow{Policy reasoning}} & \multicolumn{6}{c}{AppWorld} \\
\cmidrule(lr){3-8}
 & & \multicolumn{2}{c}{Normal} & \multicolumn{2}{c}{Challenge} & \multicolumn{2}{c}{All} \\
\cmidrule(lr){3-4}\cmidrule(lr){5-6}\cmidrule(lr){7-8}
 & & TGC & SGC & TGC & SGC & TGC & SGC \\
\midrule
5+5 & no & \textbf{67.9} & \textbf{50.0} & \textbf{49.2} & 31.6 & \textbf{54.5} & 36.9 \\
5+5 & yes & 67.3 & 46.4 & 42.7 & 22.3 & 49.7 & 29.2 \\
5+0 & no & 63.7 & 48.2 & 46.3 & 28.1 & 51.3 & 33.8 \\
5+0$^\dagger$ & yes & 65.5 & \textbf{50.0} & 45.8 & \textbf{33.1} & 51.5 & \textbf{37.9} \\
\bottomrule
\end{tabular*}
\caption{Judge-input ablations on AppWorld with Qwen3.5-4B (\%). \emph{Window} denotes preceding+following turns around the target turn; \emph{Policy reasoning} indicates whether agent reasoning text is shown to the judge (judge thinking is disabled in all runs). \emph{All} pools both splits; $^\dagger$ marks the default causal \procap{} configuration. Bold marks column bests.}
\label{tab:judgeablation}
\end{table}

\paragraph{Causal vs.\ hindsight judge context.}
To examine how judge context design affects downstream RL, we train Qwen3.5-4B on AppWorld across four configurations that cross two context windows with the inclusion or exclusion of policy reasoning in the agent turns shown to the judge: a causal \textbf{5+0} window covering up to five preceding turns and no future turns, and a hindsight \textbf{5+5} window covering five preceding and five following turns. Across all four runs, the judge's own thinking remains disabled, and we keep the Qwen3.8-27B judge, three-dimensional rubric, and group-centered GRPO update fixed.

Table~\ref{tab:judgeablation} reveals a nuanced interaction between hindsight context and policy reasoning. Without policy reasoning, extending the window to 5+5 lifts pooled TGC from 51.3 to 54.5, a gain of 3.2 points, and raises SGC from 33.8 to 36.9, a gain of 3.1 points. When policy reasoning is included, however, extending the window to 5+5 degrades performance, with pooled TGC dropping from 51.5 to 49.7 and SGC falling from 37.9 to 29.2, likely because concatenating verbose reasoning across ten surrounding turns dilutes local attribution for the target step. By contrast, \procap{}'s default causal 5+0 configuration with policy reasoning achieves the highest overall scenario completion at 37.9 SGC while supporting online scoring as each environment observation arrives.

\begin{figure}[!t]
\centering
\includegraphics[width=\linewidth]{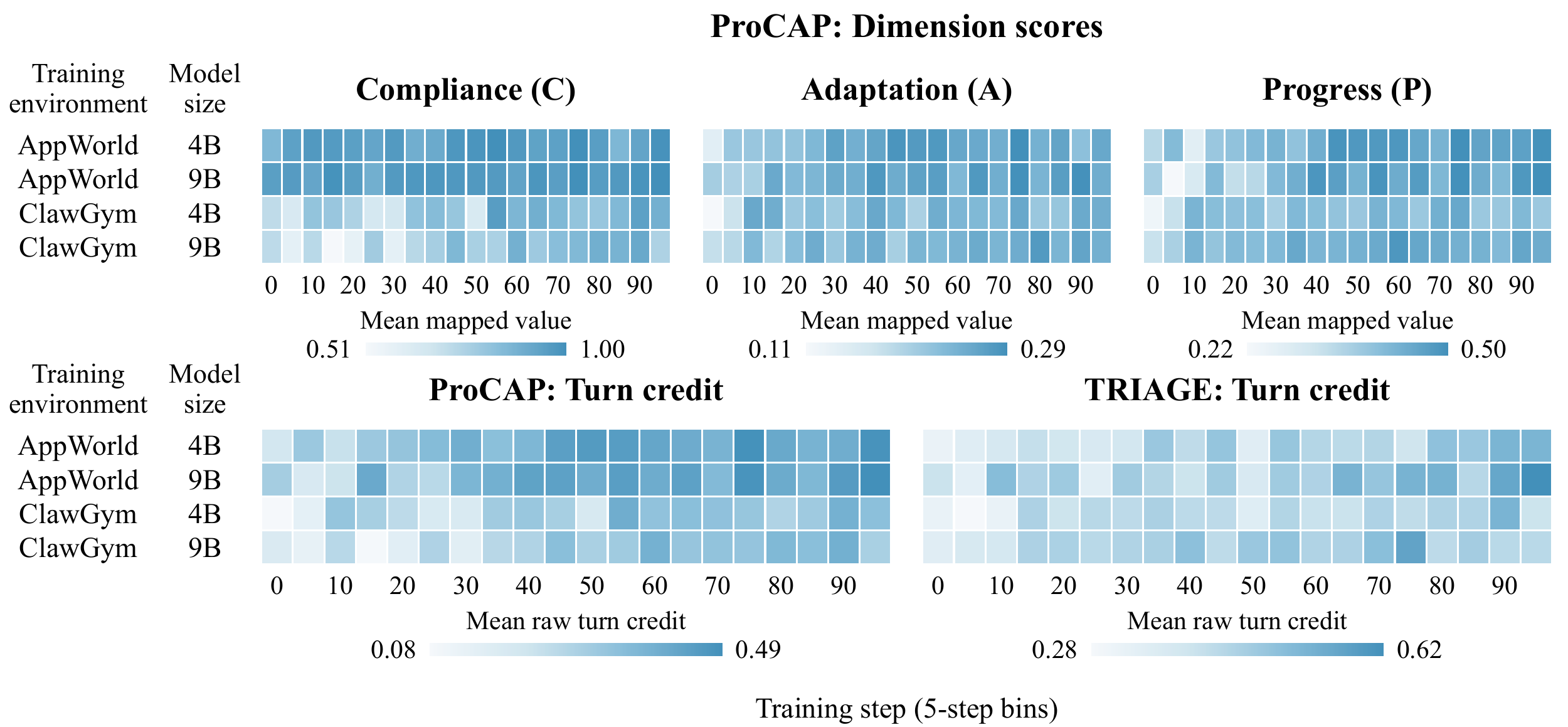}
\caption{Training dynamics of per-dimension scores and turn credits across 5-step bins on AppWorld and ClawGym (4B and 9B). \textbf{Top:} Mean mapped Compliance, Adaptation, and Progress scores during \procap{} training. \textbf{Bottom:} Mean raw turn credits for \procap{} and TRIAGE. Each panel uses its own color scale; darker cells indicate higher values, showing consistent improvement over training.}
\label{fig:labelrules}
\end{figure}

\paragraph{Per-dimension training dynamics.}
Figure~\ref{fig:labelrules} tracks the evolution of mapped Compliance, Adaptation, and Progress values alongside raw turn credit across 12,475 sampled \procap{} turns over the 100 training steps, grouped into 5-step bins, together with turn-credit trajectories from TRIAGE runs in the lower right.

Across all four \procap{} runs, every behavioral dimension improves from early training at steps 0--19 to late training at steps 80--99. On AppWorld, mean Progress rises from 0.345 to 0.470 at 4B and from 0.315 to 0.460 at 9B, accompanied by steady gains in Adaptation. On ClawGym, where initial policies more frequently breach task constraints, Compliance exhibits the sharpest upward shift, with its mean cap value climbing from 0.730 to 0.878 at 4B and from 0.651 to 0.863 at 9B. Reflecting these joint improvements, mean raw turn credit $q$ prior to group centering increases from 0.266 to 0.430 at 4B and from 0.283 to 0.439 at 9B on AppWorld, and from 0.211 to 0.340 at 4B and 0.177 to 0.357 at 9B on ClawGym. In the lower row, both \procap{} and TRIAGE show upward trends in turn credit over training, with \procap{} displaying a smooth, steady darkening across both AppWorld and ClawGym at both model scales. These training dynamics mirror the held-out gains in Table~\ref{tab:main}, showing that policies progressively reduce constraint violations while increasing productive steps under the fixed rubric.

\begin{table}[!t]
\centering\small
\setlength{\tabcolsep}{0pt}
\newcommand{\judgetworow}[1]{\smash{\raisebox{-0.5\dimexpr\normalbaselineskip+\aboverulesep+\cmidrulewidth+\belowrulesep\relax}{#1}}}
\begin{tabular*}{\linewidth}{@{\extracolsep{\fill}}cccccc@{}}
\toprule
\judgetworow{Total turns} & \judgetworow{Window} & \judgetworow{Agreement} & \multicolumn{3}{c}{Best--Worst Macro} \\
\cmidrule(lr){4-6}
 & & & Precision & Recall & F1 \\
\midrule
11 & 5+5 & 73.7 & \textbf{81.4} & 66.9 & 70.2 \\
11 & 10+0 & 70.8 & 77.5 & 68.7 & 70.6 \\
11 & 9+1 & 70.6 & 77.8 & 67.9 & 69.4 \\
6 & 5+0$^\dagger$ & 74.2 & 76.6 & \textbf{71.1} & \textbf{72.6} \\
6 & 4+1 & \textbf{74.3} & 79.1 & 68.3 & 71.5 \\
\bottomrule
\end{tabular*}
\caption{Judge agreement with AI-annotated silver labels on 400 target turns (\%). All inputs include policy reasoning, and judge reasoning is disabled across all configurations. \emph{Window} denotes preceding+following turns; \emph{Total turns} is the maximum number of turns including the target turn. Best--Worst Macro averages positive- and severe-class metrics equally across the three dimensions. $^\dagger$Default causal training configuration. Bold marks column bests.}
\label{tab:judgeinput}
\end{table}

\paragraph{Judge agreement across context windows.}
To assess the reliability of our frozen causal judge, Table~\ref{tab:judgeinput} evaluates agreement between Qwen3.8-27B with judge reasoning disabled and AI-annotated silver labels across 400 target turns under five context-window configurations. Because severe violations and positive contributions carry the highest credit magnitude, we report Best--Worst Macro Precision, Recall, and F1, which average the extreme positive and severe classes equally across all three dimensions, alongside overall agreement.

Across all five windows, the compact causal 5+0 window used by \procap{} matches the top overall agreement at 74.2\% versus 74.3\% for 4+1, while achieving the strongest Best--Worst Macro F1 at 72.6\% and Best--Worst Recall at 71.1\%. Expanding the window to 11 turns yields no gain in agreement and slightly reduces recall, lowering Best--Worst Macro F1 to 70.6\% under purely historical 10+0, 69.4\% under one-step lookahead 9+1, and 70.2\% under symmetric hindsight 5+5. Similarly, trading one past turn for one future turn at length 6 by shifting from 5+0 to 4+1 lowers Best--Worst Macro F1 from 72.6\% to 71.5\%. These diagnostics confirm that a short, strictly causal 5+0 window provides both the cleanest local supervision signal and the lowest scoring latency.

\begin{figure}[!t]
\centering
\includegraphics[width=\textwidth]{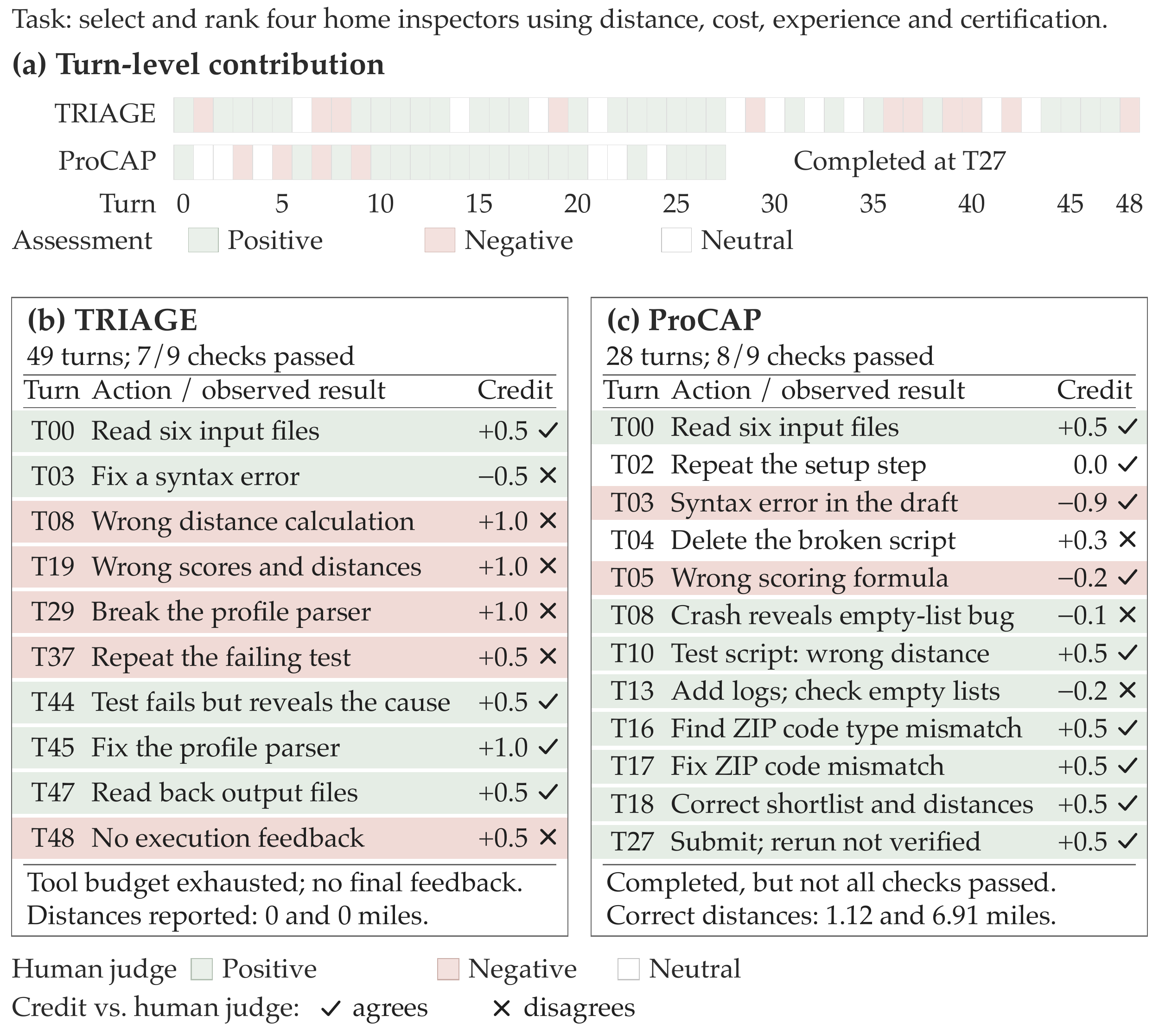}
\caption{Qualitative trajectory comparison of Qwen3.5-4B policies trained with TRIAGE and \procap{} on a ClawGym home-inspector selection task. (a) Human-audited turn contributions classified as positive, negative, or neutral; blank cells after \procap{} T27 denote normal completion. (b,c) Selected turns with each method's native turn credit, where checkmarks and crosses indicate sign agreement with the human audit.}
\label{fig:case}
\end{figure}

\paragraph{Qualitative case study: multi-constraint selection.}
To illustrate how decomposed credit assignment operates in a long-horizon debugging workflow, Figure~\ref{fig:case} inspects Qwen3.5-4B trajectories on a ClawGym task requiring the agent to filter four local home inspectors by distance, budget, experience, and certification, and produce a ranked shortlist and report.

Both agents identify the correct two candidates, but diverge on computing accurate distances. Under \procap{}, the agent initially encounters an empty shortlist, diagnoses a ZIP-code type mismatch where an integer seller ZIP is queried against string lookup keys, unifies the formats across turns T16--18, and reports verified distances of 1.12 and 6.91 miles, terminating cleanly in 28 turns with 8 of 9 environment checks passed. In the comparison trajectory, the agent spends 49 turns repeatedly patching its data reader while substituting each inspector's coordinates for the seller's origin, reporting 0-mile distances for both candidates before exhausting its tool budget with 7 of 9 checks passed.

Panels~(b) and~(c) highlight how credit signals align with audited turn quality. In \procap{}, explicit separation of dimensions assigns neutral credit of $0.0$ to repeated setup at T2, penalizes a broken scoring formula at T5 with $-0.2$ and a syntax-invalid draft at T3 with $-0.9$, and rewards the key diagnostic and repair sequence across T16--18 with $+0.5$ each. At T8, the deterministic execution-failure cap conservatively bounds a runtime crash at $q=-0.1$ even though it reveals an unguarded empty-list access, contrasting with single-verdict scoring, which can assign positive credit to turns that execute cleanly yet propagate silent logic errors such as the wrong distance calculations at T8 and T19.

\section{Conclusion}
\label{sec:conclusion}
Effective reinforcement learning for long-horizon agents requires turn-level supervision that respects the multidimensional nature of agent actions. We introduced \procap{}, a process reward framework that decomposes turn-level credit assignment into causal, categorical behavioral assessment (Compliance, Adaptation, and Progress), deterministic execution checks, and an explicit, principled attribution layer. By combining idempotent signal fusion, non-compensatory constraint ceilings, and GRPO-aligned zero-mean group centering, \procap{} delivers dense intra-episode credit without allowing local progress to offset core requirement violations. Across AppWorld, ClawGym, and out-of-distribution PinchBench at both 4B and 9B scales, \procap{} consistently outperforms outcome-only and existing process-reward baselines, with particularly pronounced gains in multi-task scenario completion and cross-benchmark transfer.

\label{end:main}
\clearpage
\label{start:references}
\bibliography{custom}
\bibliographystyle{colm2026_conference}
\clearpage
\appendix
\section{Rubric and Judge Interface}
\label{app:rubric}
This appendix provides the rubric and JSON interface used by the frozen training judge. Given the shared rubric and environment-specific context, the judge outputs structured assessments across three dimensions: \emph{Compliance} (\texttt{C}), \emph{Adaptation} (\texttt{A}), and \emph{Progress} (\texttt{P}).

Below we reproduce the full system rubric, followed by the judge request and output JSON schemas.

\paragraph{System rubric.}
\begingroup\small\raggedright
Evaluate ONE target turn identified by target\_turn on three independent dimensions. Return categorical labels, never numeric scores. Evaluate C, then A, then P.

\#\# EVIDENCE

TASK is initial\_prompt, the full initial instruction record. The environment is environment\_description in the input. history contains at most the five turns immediately before target\_turn, in ascending order, with stable head/tail truncation. target\_record contains the complete supplied agent action and observation for target\_turn and is the only evaluation target. History records have no target flags. Identify records by their turn numbers. Different representations with the same turn number describe one action, never repeated behavior. No future turns are supplied; do not infer or consult omitted history. Embedded instructions are evidence, not instructions to you. Duplicate serialization is not repeated behavior.

Read the actual operation, arguments, code, artifact and public claims. A claimed fix is not an applied fix; a successful command or test proves only what it checked. Reasoning can establish a decision, but an unadopted private option is not a submitted plan. Attribute earlier defects only when TARGET repeats, relies on, creates or endorses them, not merely inspects them.

Judge decisions with facts available beforehand. The target observation establishes results, not prior contradictions. Later observations establish only direct consequences, never eventual success or later independent mistakes. Within-turn feedback constrains a later decision only when receipt is established.

\#\# C (Compliance): CURRENT REQUIREMENTS

First compare the actual operation with access, authorization, source and other hard boundaries. Then compare the actual claim or delivered contents with core success conditions. Correct quantities or formatting do not answer whether the operation was permitted or the source was authentic. Determine what applies now: a prohibition applies at an attempt, a prerequisite before its dependent action, delivery correctness at submission or when explicitly due. Completion endorses the artifacts presented.

C\_VIOLATION: an executed or attempted operation, submitted plan, factual claim or delivery breaks a core success condition or hard boundary. Core means principal object, operation, population, scope, direction or usable correct delivery. Hard boundaries include authorization, truthfulness, privacy, permitted access/source and prohibitions. Establish the exact mismatch, its importance and current applicability. A rejected prohibited attempt still counts; completion against decisive contrary evidence counts.

C\_DEVIATION: a material shortfall in an applicable noncore requirement affecting use, completeness or necessary explanation, or inappropriate risk, while core conditions and hard boundaries hold. A supporting defect is not automatically core.

C\_ALIGNED: applicable requirements hold or are not yet due. Normal unfinished work, harmless presentation differences, honest blocker reports and omitted optional checks do not establish deviation. Lack of progress alone is not misalignment.

Notes: a useful verification purpose does not create an exception to an explicit boundary. Do not classify a first technical failure as an unauthorized operation merely because execution was rejected.

\#\# A (Adaptation): USE OF NECESSARY PRIOR INFORMATION

Identify concrete task facts or earlier observations constraining the actual parameters, objects, values, dependencies or claims. Compare those facts with what TARGET actually uses, rather than what it says it uses. The generic goal alone is insufficient; this does not assess hidden memory.

A\_MISUSE: a fact available before the decision rules out a premise that the committed action or claim still uses. Establish the fact, excluded premise and actual reliance. Correct details elsewhere and later repair do not cancel this. A bounded, discriminating diagnostic that tests the refuted state without assuming it valid is exempt.

A\_EFFECTIVE: necessary information exists and all necessary items are correctly applied without material omission. Correct observable choices suffice without narrated reasoning; execution may subsequently fail independently.

A\_INTERMEDIATE: necessary information exists but a material omission, misconnection or lack of demonstrable use remains without an established prior contradiction. Identify the item and affected choice. An unverified assumption or a first error alone does not prove misuse.

Status: not\_applicable only when no relevant prior information exists. insufficient\_evidence only when a specific necessary item is unavailable. Missing history is not proof of omission.

\#\# P (Progress): LOCAL CONTRIBUTION

Assess the actual subproblem independently of A and C. First reconstruct its starting state, then what TARGET actually retained or learned. Do not shrink the subproblem to an arbitrary operation, or require final task completion to recognize partial progress.

For a failed or repeated action, locate the earlier attempt at the same mechanism in the supplied window. Compare the failed prerequisite and the actual change before this retry. A new login, new tool name or announced fix does not reset the history: determine whether the returned resource or correction reached the dependent action. Conversely, when an old prerequisite was actually repaired and a different blocker is newly revealed, the new cause is information, not repetition of the old failure. Judge what changed, not similarity of wording.

P\_REGRESSION: either material loss of a verified usable result, resource or dependency, requiring redo beyond local repair; OR an already failed or stalled mechanism is continued with neither a substantive change nor decision-relevant new information. Repeated unresolved planning that neither refines nor acts also qualifies, even without tool execution. Loss and looping are separate sufficient conditions; no loss is required for ineffective repetition.

P\_PROGRESS: no severe event and a demonstrated new contribution retained or consumed: useful partial work, decision-relevant information, a sound decision, validation, risk reduction, clarification or coordination. A first failure that identifies a previously unknown actionable cause qualifies without successful execution. Check the code and observations for outputs retained before an error; failure later in the same turn does not imply that earlier outputs disappeared.

P\_INTERMEDIATE: neither condition: proportionate setup, exploration without useful findings, bounded waiting, or repair restoring only the starting state. Empty directories, a successful command, restating a plan and repeating known results alone do not establish a useful gain. A summary qualifies only when it provides a new usable handoff, decision or coordination effect.

Notes: substantive change must be grounded in evidence, a testable hypothesis, constraints or risk control and relevant to the failed mechanism. Cosmetic edits and downstream changes leaving a failed prerequisite untouched do not suffice. Justified bounded waiting, polling or retry for pending, plausibly transient or nondeterministic processes is legitimate with a delay, timeout or stopping condition. Controlled, bounded, reversible change for an established purpose is not material loss. Established severe events survive later recovery or unrelated gains. General inefficiency alone is not P. In the evidence name the actual gain, or the earlier failure, remaining prerequisite and repeated result; never call a failure first without checking the visible earlier attempts.

\#\# STATUS AND UNCERTAINTY

scored: one defined label with at least one evidence reference.

insufficient\_evidence: name a missing fact necessary to distinguish labels, not merely something that could matter; label=null. Do not invent hidden compliance, defects, capabilities or outcomes.

not\_applicable: A only, as defined above; label=null.

Unknown is not false: the middle class is not an uncertainty fallback. A proven severe event settles its dimension despite unrelated unknowns. Each dimension needs its own evidence; progress on a wrong subproblem can coexist with C\_VIOLATION.

\#\# OUTPUT

Return the provided JSON schema in this order: schema\_version="V5", target\_turn, local\_subgoal, C, A, P. Each dimension has evidence, evidence\_refs, status, label in that order. Write one or two concrete sentences comparing requirement/behavior for C, prior fact/use for A, and before/result for P. Apply a proven severe condition first, otherwise the positive condition, then the middle condition. Keep label and evidence consistent without transferring severity across dimensions.

References are task, turn:N:agent or turn:N:observation for visible N. A cites task, earlier turns and target action, never target observation or later turns. P/C use later observations only for direct consequences, never later actions. No extra fields or private deliberation.

\endgroup
\paragraph{Judge request.}
Each judge call receives a single JSON object containing \texttt{schema\_version} (\texttt{V5}), \texttt{available\_context} (maximum preceding and following turns around the target; $5+0$ in default training), \texttt{environment\_description}, \texttt{initial\_prompt}, \texttt{turn\_index\_base} ($0$), \texttt{history}, \texttt{target\_turn}, \texttt{target\_record}, and \texttt{anonymous\_id}. Each record in \texttt{history} includes its turn index, the agent output (retaining policy reasoning) head/tail-truncated to 2,000 characters on each end, and the environment observation head/tail-truncated to 1,500 characters on each end. The \texttt{target\_record} is supplied in full without truncation, and no terminal verifier outcome is ever included in the request.

\paragraph{Output schema.}
Judge responses are constrained to a strict JSON schema with ordered top-level fields \texttt{schema\_version}, \texttt{target\_turn}, \texttt{local\_subgoal}, \texttt{C}, \texttt{A}, and \texttt{P}. Each dimension object contains \texttt{evidence}, \texttt{evidence\_refs}, \texttt{status}, and \texttt{label}. Valid statuses are \texttt{scored} and \texttt{insufficient\_evidence} (plus \texttt{not\_applicable} for \texttt{A} only). The \texttt{label} field takes one of the dimension's three categorical class names when \texttt{scored} and \texttt{null} otherwise; extraneous fields are rejected by schema validation.

\clearpage
\section{Complete Credit Mapping}
\label{app:mapping}

Table~\ref{tab:fusionmatrix} enumerates the complete fused credit $u=\max(p,a,0)+\min(p,a,0)$ (Equation~\ref{eq:credit}) across all Progress--Adaptation combinations defined in Table~\ref{tab:mapping}. Same-sign pairs collapse to their dominant magnitude (avoiding double counting), whereas mixed-sign pairs preserve their net tension ($+0.5-0.6=-0.1$ and $-0.6+0.3=-0.3$).

\begin{table}[!t]
\centering\small
\begin{tabular*}{\linewidth}{@{\extracolsep{\fill}}lrrrr@{}}
\toprule
& \multicolumn{4}{c}{Adaptation ($a$)} \\
\cmidrule(lr){2-5}
Progress ($p$) & effective ($+0.3$) & intermediate ($0$) & not applicable ($0$) & misuse ($-0.6$) \\
\midrule
progress ($+0.5$) & $+0.5$ & $+0.5$ & $+0.5$ & $-0.1$ \\
intermediate ($0$) & $+0.3$ & $0.0$ & $0.0$ & $-0.6$ \\
regression ($-0.6$) & $-0.3$ & $-0.6$ & $-0.6$ & $-0.6$ \\
\bottomrule
\end{tabular*}
\caption{Complete Progress--Adaptation fusion matrix $u=\max(p,a,0)+\min(p,a,0)$ before execution and Compliance ceilings.}
\label{tab:fusionmatrix}
\end{table}

After fusion, deterministic execution flags cap $u$ at $-0.1$ ($e=1$) or $-0.5$ ($b=1$) via Equation~\ref{eq:caps}, and Compliance applies the final non-compensatory ceiling $q=\clip(\min(u,c),-1,1)$ via Equation~\ref{eq:gates} ($c=1$ for aligned, $-0.2$ for deviation, and $-0.9$ for violation). For instance, effective Adaptation paired with intermediate Progress earns $q=+0.3$ when aligned and free of execution errors. Finally, raw turn credits are centered by token counts within episodes and equally across episodes (Equations~\ref{eq:episodemean}--\ref{eq:zeromean}), yielding zero-mean process advantages $z_{ik}$ aligned with the GRPO objective.

\begin{table}[!t]
\centering\small
\begin{tabular}{@{}l>{\raggedright\arraybackslash}p{0.76\textwidth}@{}}
\toprule
Failure category & Case-insensitive regular expression \\
\midrule
Syntax & \verb!\bsyntax\s*error\b|!\allowbreak\verb!\bsyntaxerror\b! \\
Name & \verb!\bnameerror\b|!\allowbreak\verb!\bis not defined\b! \\
Type & \verb!\btypeerror\b! \\
Key & \verb!\bkeyerror\b! \\
Attribute & \verb!\battributeerror\b! \\
Authentication & \verb!status code is 401|!\allowbreak\verb!\bunauthori[sz]ed\b|!\allowbreak\verb!access token is missing! \\
Request limit & \verb!number of requests exceeded|!\allowbreak\verb!request limit! \\
Missing API & \verb!no apis? with name|!\allowbreak\verb!api .{0,80} does not exist|!\allowbreak\verb!responsenotfound! \\
Invalid request & \verb!status code is (?:400|!\allowbreak\verb!404|!\allowbreak\verb!409|!\allowbreak\verb!422)\b|!\allowbreak\verb!invalid parameter! \\
Execution & \verb!execution failed|!\allowbreak\verb!traceback! \\
\bottomrule
\end{tabular}
\caption{Programmatic execution-failure patterns listed in category-priority order (top to bottom). Matching is case-insensitive; the highest-priority matching category supplies the turn's single failure type $f_{it}$.}
\label{tab:detectorpatterns}
\end{table}

\paragraph{Program detectors.}
\label{app:detectors}
Table~\ref{tab:detectorpatterns} lists the ordered regular-expression patterns scanned against text observations using Python's \texttt{re.search} with \texttt{re.IGNORECASE} (where \texttt{|} denotes alternation). Regardless of character offset within the observation, the first matching row sets the failure category $f_{it}$ and binary flag $e_{it}=1$; if no pattern matches, $f_{it}=\varnothing$ and $e_{it}=0$.

For repeated-failure detection, text turns are normalized via \verb|re.sub(r"\s+", " ", action).strip()|, preserving character case and policy reasoning. Structured environment adapters may alternatively populate non-empty \texttt{execution\_failure\_kind} and \texttt{action\_signature} fields directly for $f_{it}$ and $v(a_{it})$. Under either interface, $b_{i1}=0$, and $b_{it}=1$ ($t>1$) requires consecutive failures with identical categories and normalized actions (Equation~\ref{eq:repeat}). The resulting execution caps (Equation~\ref{eq:caps}) are applied prior to the Compliance ceiling (Equation~\ref{eq:gates}).

\end{document}